\documentclass[a4paper, 10pt, conference]{ieeeconf}      
\usepackage{FG2025}
\usepackage{amsmath,amsfonts}
\usepackage{graphics}
\usepackage{algorithmic}
\usepackage{algorithm}
\usepackage{array}
\usepackage[caption=false,font=normalsize,labelfont=sf,textfont=sf]{subfig}
\usepackage{textcomp}
\usepackage{stfloats}
\usepackage{url}
\usepackage{verbatim}
\usepackage{graphicx}
\usepackage{cite}
\usepackage{booktabs}
\usepackage{amsmath}                 
\usepackage{amssymb}
\usepackage{balance}
\usepackage{xcolor}

\FGfinalcopy 

\IEEEoverridecommandlockouts                              
\def\FGPaperID{42} 

\title{\LARGE \bf
Dimitra: Audio-driven Diffusion model for 

Expressive Talking Head Generation}

\author{\parbox{16cm}{\centering
    {\large Baptiste Chopin$^1$ Tashvik Dhamija$^1$ Pranav Balaji$^1$ Yaohui Wang$^2$ Antitza Dantcheva$^1$}\\
    {\normalsize
    $^1$ Université Côte d’Azur, Inria, STARS Team, France\\
    $^2$ Shanghai Artificial Intelligence Laboratory, China}}
}

\begin{document}

\ifFGfinal
\thispagestyle{empty}
\pagestyle{empty}
\else
\author{Anonymous FG2025 submission\\ Paper ID \FGPaperID \\}
\pagestyle{plain}
\fi
\maketitle

\begin{abstract}

 We propose Dimitra, a novel framework for audio-driven talking head generation, streamlined to learn lip motion, facial expression, as well as head pose motion. Specifically, we train a conditional Motion Diffusion Transformer (cMDT) by modeling facial motion sequences with 3D representation. We condition the cMDT with only two input signals, an audio-sequence, as well as a reference facial image. By extracting additional features directly from audio, Dimitra is able to increase quality and realism of generated videos. In particular, phoneme sequences contribute to the realism of lip motion, whereas text transcript to facial expression and head pose realism. 
Quantitative and qualitative experiments on two widely employed datasets, VoxCeleb2 and HDTF, showcase that Dimitra is able to outperform existing approaches for generating realistic talking heads imparting lip motion, facial expression, and head pose. 
\end{abstract}

\section{Introduction}
\label{sec:intro}

Talking head generation aims at animating face images, placing emphasis on generation of realistic appearance and motion. The latter has been enabled by the rapid progress of generative models. Related results have sparked attention in domains of application including digital humans, AR/VR, as well as film-making. While \textit{video-driven} talking head generation has become highly realistic~\cite{siarohin2019first,wang2022latent,zhao2022thin}, animation driven by \textit{audio-speech} allows for additional applications such as video games and chat-bots. 
Audio-driven talking head generation models \cite{hong2022depth,li2024ae,wav2lips} entail the animation of a face image by synchronizing the audio-speech to lip motion.  
Hence, related work~\cite{wav2lips,guan2023stylesync} predominantly focuses on generating \textit{lip motion}.  
However, it is only when \textit{head pose} and \textit{facial expressions} are animated that talking heads appear realistic, as such facial behavior is crucial in \textit{human communication}. 
Motivated by this, most recent methods~\cite{ma2023styletalk,ye2024real3d} attempt to incorporate such facial behavior. 
Such methods have primarily utilized \textit{additional existing video sequences}, in order to condition the generation of \textit{facial expression}. \textit{W.r.t.} head pose, related movement has been mainly directly copied and transferred from real video sequences, which might initially appear realistic. However this is limited, as head pose sequences might be of different length, and more importantly, will not be in accordance with the speech to the generated video, resulting in unnatural motion. 

\noindent Deviating from the above, in this work, we introduce Dimitra, a novel framework for audio-driven talking head generation, streamlined to animate a face image \textit{locally and globally} based on audio speech. Specifically, we place emphasis on generating natural and diverse face motion and appearance by \textit{learning} intrinsically \textit{behavior of talking faces} that includes motion of lips, head pose, as well as facial expressions - directly \textit{from an audio input}. Towards this, we propose a conditional Motion Diffusion Transformer (cMDT), which accepts a reference facial image, as well as an audio sequence as inputs. The latter 
contributes to (i) Wav2Vec \cite{schneider2019wav2vec} features, (ii) text transcript of the audio-speech, as well as (iii) phoneme sequences that we then employ as input of the following network. In particular, the latter utilizes an intermediate 3D mesh  representation that facilitates generation of facial motion, namely 3DMM \cite{deng2019accurate}. This is beneficial in reducing the number of parameters of Dimitra and improving head motion in the 3D space. In addition, 3DMMs allow for flexibility \textit{w.r.t.} image resolution, as we are able to simply alter the final video renderer.

\noindent Our main contributions include the following.
\begin{itemize}
    \item We introduce a novel audio-driven talking head generation model, referred to as Dimitra, which generates motion pertained to lips, expression, as well as head pose in a \textit{reference facial image} based on a \textit{single audio sequence}. We extract multiple features from the audio sequence, conditioning the generation of talking head videos. Deviating from previous methods, we locally animate the mouth, as well as globally the entire face by generating facial expressions and head pose - without additional inputs. Such facial behavior is merely extracted and \textit{learned from audio sequences}. 
    \item We conducted extensive experiments that quantitatively and qualitatively demonstrate that videos generated by Dimitra are realistic, imparting expressive and natural motion.
\end{itemize}

\section{Related Work}
\label{sec:Related}



\textbf{Audio-driven Talking head generation} methods can be \textit{person specific} \cite{ji2021audio-driven,textbasedediting}, where videos can only be generated of persons that have been in the training set, clearly limiting the generation setting. More related to our problem, \textit{person agnostic} methods are able to animate unknown identities in RGB \cite{Zhou2021Pose,Wang_2023_CVPR}, neural radiance fields \cite{li2024ae}, facial landmarks in a 3D space \cite{gururani2022SPACE,wang2021audio2head}, as well as mesh representations \cite{ma2023styletalk,zhang2023sadtalker} employing numerous architectures such as LSTM \cite{10.1145/3414685.3417774}, CNN \cite{wav2lips} or diffusion \cite{wang2024eat,yu2023talking,shen2023difftalk}. It is worth noting that there exist methods that animate 3D meshes \cite{sun2024diffposetalk}, deviating from our work, as our framework Dimitra provides an RGB video as output. Originally, methods focused on generating merely lip motion, render generated videos rather unrealistic \cite{wav2lips}. More recently, research has focused on generating motion pertaining to the entire face, including facial expression and head pose, however harnessing such directly or as strong condition from real video sequences, which are required as additional inputs \cite{ma2023styletalk,ma2023dreamtalk}. We note that such methods provide good results, in case that expression and head pose sequences are manually selected. 

\textbf{Diffusion Models}~\cite{pmlr-v37-sohl-dickstein15,ho2020denoising} have shown remarkable results in several tasks including image generation~\cite{ho2020denoising,zhou2023shifted} as well as video generation \cite{harvey2022flexible,wu2023tune}. Related to our setting, diffusion models have been proposed towards generation of face images \cite{huang2023collaborative,kim2023dcface}, as well as of talking heads \cite{stypulkowski2024diffused,du2023dae}. However, existing methods have not generated head pose or facial expression, while being dataset specific. We here propose a diffusion model, designed to generate videos of talking heads, endowed with local lip motion and facial expression, as well as global head pose, animating facial images of identities beyond the training set based on an audio-speech-inputs. 

\begin{figure*}[!t]
  \centering

   \includegraphics[width=0.8\linewidth]{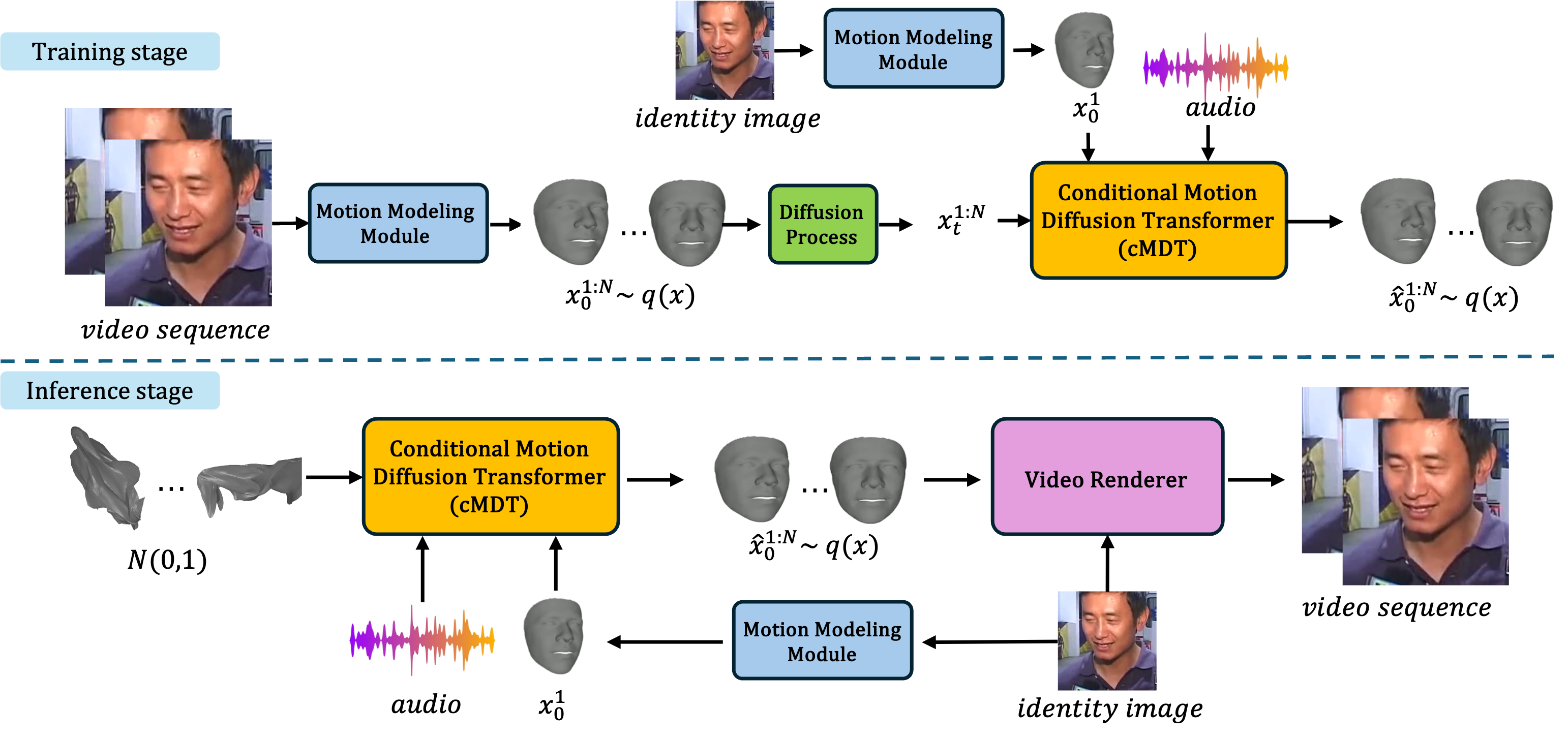}
   \caption{\textbf{Dimitra pipeline.} Dimitra comprises  three main parts, a Motion Modeling Module (MMM), a Conditional Motion Diffusion Transformer (cMDT) and a Video Renderer. In the training stage, 3D meshes (3DMM) are extracted from a video by the MMM. They are used by the cMDT jointly with features extracted from an audio sequence, to noise then denoise the 3DMM sequence.
   In the inference stage, using an audio sequence and an identity 3DMM as condition, cMDT aims at generating a 3DMM sequence from Gaussian noise. Finally, the Video Renderer transforms the 3DMM sequence into a RGB video.}
   \label{fig:dimitra}
\end{figure*}

\section{Method}
\label{method}

In this work, our goal is to generate RGB videos, where a reference facial image has been animated based on a provided audio-speech input. 
Towards this, we introduce Dimitra, comprising of three parts (see Fig.~\ref{fig:dimitra}), namely a Motion Modeling Module (MMM), a Conditional Motion Diffusion Transformer (cMDT) and a Video Renderer. While the MMM extracts motion features from the RGB image using 3D mesh, the cMDT extracts relevant features from audio that jointly, with the 3D mesh, are targeted to generate a 3D mesh sequence of facial motion. Finally, the Video Renderer converts the meshes into the RGB space, in order to create an RGB video. We proceed to elaborate below.

\subsection{Motion Modeling Module}
\label{MMM}
While we seek to generate RGB videos, recent work \cite{zhang2023sadtalker} has showcased that generating mesh representations allows for increased 
control of generated motion in the 3D space that contributes to realistic generation results that are highly limiting parameters in the network. 
Motivated by this, we here adopt a facial mesh representation, namely the 3DMM representation proposed by Deng \emph{et al.} \cite{deng2019accurate}.
Firstly, we encode the RGB data into 3D meshes by adopting an encoder \cite{deng2019accurate}. 
The parameters encode texture, identity, luminosity,  and importantly 64 parameters, encoding facial landmarks and 6 parameters representing global head pose (rotation and translation). In the following, we only refer to these 64+6 parameters as facial features. We note that out of the 64 facial parameters, 13 encode the mouth region.
Formally, from a video, we obtain a 3DMM sequence $x^{1:N}={x^1,...,x^N}$ with $x^i \in \mathbb{R}^{d}$ denoting a single 3DMM frame containing $d$ motion parameters and $N$ frames in the sequence.
\subsection{Conditional Motion Diffusion Transformer (cMDT)}

We formulate talking head generation as a reverse diffusion process and propose a Conditional Motion Diffusion Transformer (cMDT) to learn this process. The cMDT takes an audio sequence and the 3DMM representation of a face as an input (Sec. \ref{MMM}) towards generating a 3DMM sequence of facial motion. The cMDT can be divided into Audio Feature Extractor, Motion Transformer and Video Renderer, three parts.
\subsubsection{Diffusion background}

Denoising diffusion probabilistic models (DDPMs)~\cite{sohl2015deep} 
entail two processes, namely forward diffusion, as well as reverse diffusion. During the forward diffusion process, Gaussian noise is gradually added to the data up to the point, the data becomes Gaussian noise. On the other hand, during the reverse process, a neural model learns to gradually denoise the data.

\subsubsection{Audio Feature Extractor}
Towards employing the audio sequence as condition, we extract relevant features from raw audio. 
To leverage the strengths of both, phonemes representation and audio features, we proceed to employ both, \textit{audio features} Wav2Vec \cite{schneider2019wav2vec} features, \textit{as well as phonemes} as input of our model. Wav2Vec is selected for associated efficiency at encoding speech. In addition, we utilize a \textit{text transcript}, in order to obtain semantic information. We employ a phoneme aligner \cite{yuan2008speaker}, providing phonemes with timestamps from an audio-input and the corresponding text transcript. Based on the timestamps we create a sequence of phonemes, corresponding to the frame rate of the video and tokenize it. We note that phonemes constitute the smallest discrete speech units and contain essential information for word articulation. Formally we obtain $a^{1:N}={a^1,...,a^N}$ a Wav2Vec feature sequence with $a_i\in \mathbb{N}$ being the features in one frame; a tokenized phoneme sequence $p^{1:N}={p^1,...,p^N}$ with $p_i\in \mathbb{N}$ being a single phoneme token and a text transcript $S$.

\subsubsection{Motion Transformer}

We propose a Transformer architecture that accepts as input a Wav2Vec feature sequence $a^{1:N}$, a tokenized phoneme sequence $p^{1:N}$, a 3DMM sequence $x^{1:N}$, and the text transcript corresponding to the audio $S$. Additionally, we provide the first frame of the 3DMM sequence $x^1_0$ to condition the network on the first pose of the sequence to ensure that the generated 3DMM sequence starts from the original position. We formulate the facial motion generation as a reverse diffusion problem, where we sample a random noise $x^{1:N}_T$ towards obtaining a real 3DMM, representing the facial motion corresponding to the inputs. We propose in this context a Transformer architecture to learn the denoising process.
In this architecture, $S$ is encoded using a pretrained CLIP \cite{clip} Transformer encoder, $a^{1:N}$ and $p^{1:N}$ are encoded using simple trainable Transformer encoders and $x^1_0$ is encoded using a simple MLP encoder. $x^{1:N}_t$ is then denoised using a Transformer decoder containing self and cross attention layers to learn the relationship between the motion and the conditions. 
Towards reducing complexity, all attention layers employ efficient attention \cite{shen2021efficient}.



\subsubsection{Learning}
\label{losses}
Best results in our experiments are obtained for Dimitra, in case that we train separate models for lip motion, facial expression and head pose. This is partly due to the unbalanced number of parameters between each component (13 for lips, 51 for face, 6 for head pose). Consequently, we train three separate models with the same network architecture and the same losses except for the head pose model. In these, for $x^i \in \mathbb{R}^{d}$, $d=13$, $d=51$ and $d=6$ for the lip model, facial expression model and head pose model, respectively.
We only utilize the weighted diffusion loss $L_{G} = \lambda \times L_{diff}$ as training loss for the lips and expression models. Head pose, however is very sensitive to the previous pose in the sequence, and discontinuities in the generation are highly noticeable, in particular when generating recursively, in order to obtain long sequences. To limit these discontinuities, in addition to giving the first pose as input, we add a first pose diffusion loss that corresponds to diffusion only on the first frame $x^1_t$. Consequently, the loss for the head pose model is $L_{G} = \lambda \times L_{diff} + L_{first}$. $\lambda=6$ in our implementation. 
\subsection{Video Renderer}
To transfer the generated 3DMM data to video space, we utilize a pretrained image renderer, proposed by Ren \textit{et al.}  \cite{9711291} that is able to generate videos with $256\times256$ resolution from a 3DMM input. Nevertheless, 3DMM data is not limited to this resolution, rather depends on the renderer. 
Furthermore, 3DMM is transferable to other 3D mesh formats, enabling versatility in applications.

\section{Experiments}
\label{sec:Experiments}

\subsection{Experimental settings}

We train and test our network on a subset of VoxCeleb2 \cite{Chung18b}. For testing only, we also use the HDTF dataset \cite{zhang2021flow}.  We evaluate Dimitra by computing standard metrics for talking head generation. 
Specifically, F-LMD and M-LMD \cite{chen2018lip,ma2023styletalk} evaluate the facial and mouth landmarks distance, respectively, whereas we employ the SyncNet \cite{chung2017out} distance and confidence score towards evaluating lip synchronisation of generated video and audio. We compare our framework to state of the art methods StyleTalker \cite{ma2023styletalk}, SadTalker \cite{zhang2023sadtalker}, and DreamTalk \cite{ma2023dreamtalk}. \textbf{More details about datasets, metrics and baselines can be found in the supplementary material (SM).}

\begin{table}[!t] 
	\centering
		\caption{Quantitative results pertained to the VoxCeleb2 dataset}
		\resizebox{0.9\linewidth}{!}{%
\begin{tabular}{@{}c |c c lc c c cl} 
\toprule
Method                                      & F-LMD$\downarrow$       & M-LMD$\downarrow$  & $sync_{dist}\downarrow$ &$sync_{conf}\uparrow$\\
\midrule
\textbf{GT}                                        & -       & -       &8.08           &5.80            \\
\midrule

StyleTalker \cite{ma2023styletalk}        & \underline{5.73}& \underline{4.41} &11.16  &3.24 \\
SadTalker \cite{zhang2023sadtalker}       & 6.14& 4.70 &\underline{9.12}   &\underline{5.28} \\

DreamTalk \cite{ma2023dreamtalk}          & 5.78& 4.43 &\textbf{8.48}   &\textbf{5.64}\\

 Dimitra & \textbf{5.00}& \textbf{3.79}& 9.43  & 4.93\\ 
 \hline
\end{tabular}}
\label{tab:results_vox}
\end{table}

\subsection{Quantitative results}
\label{sec:quanti}
Tables \ref{tab:results_vox} and \ref{tab:results_HDTF} depict quantitative results pertained to the datasets VoxCeleb2 and HDTF, respectively. We observe that our method outperforms or is competitive \textit{w.r.t.} the state of the art on both datasets. 
Regarding \textit{landmark based metrics (F-LMD, M-LMD)} our method outperforms the other methods on both datasets. This indicates that given an audio sequence, Dimitra generates realistic videos, resembling the ground truth, as opposed to other methods \emph{w.r.t.} lip motion and facial expression. 
These results is supported by the \textbf{user study in our SM}. 

Dimitra is competitive \textit{w.r.t.} \textit{lip synchronisation metrics}. 
While qualitative results and the related user study (see SM) demonstrate the realism and synchronization between lip movement and audio-speech, the 
lip synchronisation (\textit{i.e.,} syncnet) results do not reflect on that. This stems from the fact that syncnet-metrics are not adequate in evaluating these. 
For example, SadTalker obtains competitive results \emph{w.r.t.} syncnet distance and confidence, while being able to only generate two states for the mouth, namely open and closed. Intermediate mouth shapes corresponding to pronounced sounds are not well generated (Fig.\ref{fig:example_vox1}). 

We note that deviating from DreamTalk, our framework Dimitra was not trained on HDTF. 

In Table \ref{tab:results_HDTF} we show results of two versions of our method: with and without head pose (Dimitra (HP) and Dimitra (no HP) respectively). For a single 3DMM generated by the cMDT we use one directly (HP)  while we freeze the head pose 3DMM parameters at the renderer level for the other (no HP). 
Surprisingly, while the F-LMD and M-LMD improve, the syncnet score decreases. This should not be the case, as only head pose was changed, while lip motion and expression remain exactly the same. This indicates that the metrics are instable.

\begin{table}[!t] 
	\centering
		\caption{Quantitative results pertained to the HDTF dataset}
		\resizebox{0.9\linewidth}{!}{%
\begin{tabular}{@{}c |c c lc c c cl} 
\toprule
Method                                      & F-LMD$\downarrow$       & M-LMD$\downarrow$  & $sync_{dist}\downarrow$ &$sync_{conf}\uparrow$&\\
\midrule
\textbf{GT}                                          & -     & -       &8.63   &6.82        \\
\midrule


StyleTalker\cite{ma2023styletalk}             & \textbf{2.51}&  \underline{2.42}&13.06  &1.88      \\
SadTalker \cite{zhang2023sadtalker}         & 3.74&  4.19 &10.36  &5.34     \\
\
DreamTalk\cite{ma2023dreamtalk}             & 2.52&  \underline{2.42} &\textbf{8.66}   &\textbf{6.14}      \\

Dimitra (HP)& 3.85&  3.23&\underline{9.42}   &\underline{5.69}    \\ 

 Dimitra (no HP)& \textbf{2.51}& \textbf{2.38} & 9.60& 5.49 \\
\hline
\end{tabular}}
\label{tab:results_HDTF}
\vspace{-0.5cm}
\end{table}

\subsection{Qualitative results}\label{sec:qual_res}

\begin{figure*}[t]
  \centering

   \includegraphics[width=1.0\linewidth]{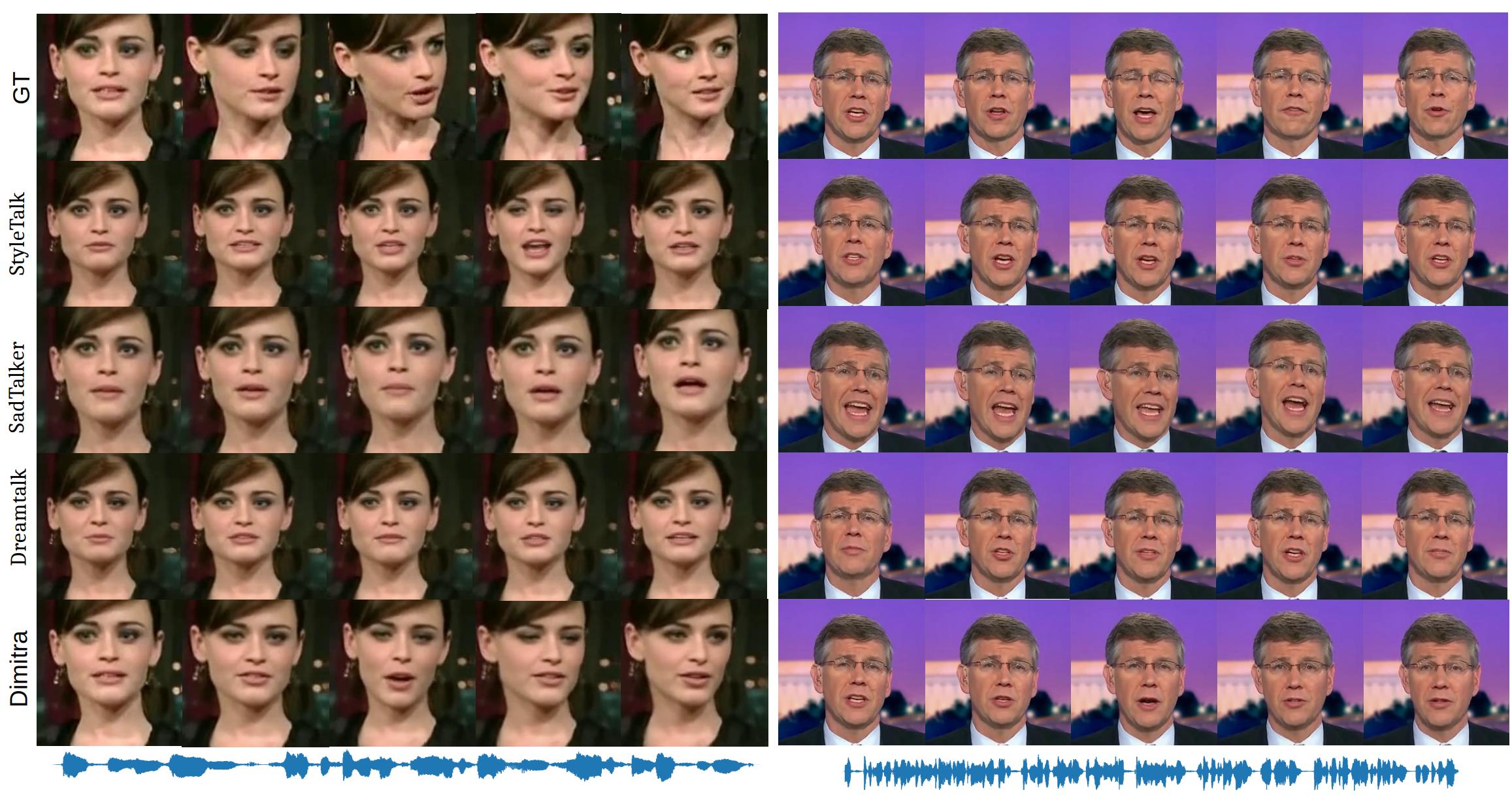}
   \caption{\textbf{Qualitative results.} Examples of generated samples pertained to the VoxCeleb2 dataset on the left and pertained to HDTF on the right.}

   \label{fig:example_vox1}

\end{figure*}

\noindent We present qualitative results pertained to the VoxCeleb2 and HDTF datasets in Figure \ref{fig:example_vox1}. \noindent Specifically, our method generates coherent videos that  synchronize well with the ground truth, temporally and spatially. Notably, results \emph{w.r.t.} HDTF are convincing, despite our model not being trained on this dataset. All other methods comprise visual spatio-temporal synchronization errors. 
We also showcase that as discussed in Sec.\ref{sec:quanti}, SadTalker does not generate continuous mouth shapes, it only generates an opened or closed mouth, sometimes failing to generate a coherent mouth (Figure \ref{fig:example_vox1}, right side). Despite these issues, SadTalker obtains high lip-sync scores in the quantitative analysis, indicating the limitation of the metric.

We observe that our method generates expressive talking faces. While DreamTalk  includes facial expressions in generated videos, they require a strong conditioning based on an additional real "style sequence". Dimitra is the \textit{only method to generate realistic head pose motion}. 

\section{Conclusions}

We introduced Dimitra, a network streamlined to generate talking head videos based on a reference facial image and an audio-speech sequence. Deviating from previous models, Dimitra endows the reference image with lip motion associated to the audio, as well as with learned facial expression and head pose. Our qualitative and quantitative results showcase superiority \emph{w.r.t.} state of the art on two widely used datasets, due to reliable lip synchronization and expressiveness in our generated videos. By extracting features (phoneme and text transcript) directly from audio-speech, Dimitra generates local (\textit{i.e.,} facial expression) and global (\textit{i.e.,} head pose) motion.
Dimitra is able to generate realistic videos, animating subjects outside of the original training data distribution. Future work will focus on learning representations and generating the entire upper body including articulated hand gestures. The code, preprocessing code and training - testing split will be released. 

\clearpage


{\small
\balance
\bibliographystyle{ieee}
\bibliography{main}
}

\pagebreak
\begin{center}
\textbf{\large Dimitra: Audio-driven Diffusion model for Expressive Talking Head Generation Supplemental Material}
\end{center}
\setcounter{equation}{0}
\setcounter{figure}{0}
\setcounter{table}{0}
\setcounter{page}{1}
\setcounter{section}{0}
\renewcommand{\thesection}{S-\Roman{section}}
\makeatletter
\renewcommand{\theequation}{S\arabic{equation}}
\renewcommand{\thefigure}{S\arabic{figure}}

In this supplementary material we provide additional details \textit{w.r.t.} user study, as well as datasets, baselines and metrics. 

\section {User study}
\label{sec:userstudy}
We conduct a user study to evaluate generated video quality. Towards achieving fair evaluation, we display paired videos or generated by the 4 approaches (Dimitra, Styletalker, Sadtalker, Dreamtalk) on the HDTF dataset and ask 20 human raters for each paired video the question `which clip is more realistic and natural?'.
Each video-pair contains a generated video from our method, as well as a video generated from Styletalk, Sadtalker, Dreamtalk or the ground truth.

\begin{table}[!thb]
\caption{\textbf{User study.} We ask 20 human raters to conduct a subjective video quality evaluation on the HDTF dataset. Results show that videos generated by Dimitra are consistently rated as more realistic.}
\label{tab:user_study}
\begin{center}
\setlength{\tabcolsep}{3.3pt}
\setlength\arrayrulewidth{1pt}
\begin{tabular}{cccc}
\hline
& user preference (\%)\\
\hline
Dimitra/Styletalk   & \textbf{87.5}/12.5 \\
Dimitra/Sadtalker   & \textbf{75.0}/25.0 \\
Dimitra/Dreamtalk   & \textbf{68.8}/31.2 \\
\hline
\end{tabular}
\end{center}
\end{table}

Results suggest that our generated videos are the most realistic in comparison to other methods on the HDTF dataset, see Table~\ref{tab:user_study}. 





\section{Datasets, Baselines and Evaluation metrics}
\subsection{Datasets}
\noindent\textbf{VoxCeleb2} is a dataset containing more than 1 millions low resolution clips (average length 10s) of people talking in TV shows or interviews. We only use a subset of 50000 clips of which 3931 are used for testing (corresponding to 39 identities). We use face crop with a resolution of 256*256 but the quality of face crops vary as faces sizes in original videos can be widely different. The data is diverse with many different settings, a lot of facial expressions and head motions.

\noindent\textbf{HDTF} is a datasets containing 400 videos (average length 3-5 min) of people talking, all videos are using for testing. The video resolutions vary between 512*512 and 1024*1024, we resize all videos to 256*256 for a fair comparison with the other methods. Unlike VoxCeleb2, HDTF doesn't contain many facial expression or head motions. 

\subsection{Baselines}
We compare our framework to state of the art methods StyleTalker \cite{ma2023styletalk}, SadTalker \cite{zhang2023sadtalker}, DreamTalk \cite{ma2023dreamtalk}. We note that StyleTalker and DreamTalk generate a video based on an input facial image and an audio sequence by creating a 3DMM representation similarly to Dimitra. Deviating from our model, StyleTalker and DreamTalk additionally require a head pose sequence, as well as a "style sequence" that captures the identity of the speaker and speaking style. For the sake of fairness, in the scope of generating talking heads only from audio and one image, we provide StyleTalker and DreamTalk with each first frame of the 3DMM sequence as "style sequence" and "head pose sequence". 
StyleTalker only employs a phoneme representation for the audio, whereas DreamTalk incorporates Wav2vec features \cite{schneider2019wav2vec}. As the StyleTalker-code associated to their phoneme extractor is not publicly available, we convert our phonemes to their format as input. SadTalker accepts audio and image as inputs towards generating talking head endowed with head pose motion. The training code for StyleTalker, DreamTalk and SadTalker were not available, therefore we use pretrained weights, as provided by the authors. These were obtained by training on a number of datasets including HDTF and VoxCeleb2. Hence, samples of our training set might have been part of the training set of these methods as well.

\subsection{Metrics}
We evaluate Dimitra by computing standard metrics for talking head generation. 
Specifically, F-LMD and M-LMD \cite{chen2018lip,ma2023styletalk} evaluate the facial and mouth landmarks, respectively (68 and 20 landmarks extracted with dlib \cite{dlib09}, respectively). Both metrics compare landmarks pertained to generated sample and ground truth, and compute the average Euclidean norm considering landmarks and frames. We normalize the landmarks \emph{w.r.t.} the head pose when evaluating F-LMD and M-LMD using the Kabsch-Umeyama algorithm \cite{88573} between the ground truth landmarks and the generated landmarks. This is done independently for each frame of the sequence and allow us to remove the influence of head pose translation and rotation when evaluating the expression and lips motion. It also scale the data which is especially important when evaluating Dreamtalk and Styletalk on VoxCeleb2 as the methods change the viewpoint. Due to the large variations in head pose in the VoxCeleb2 dataset, dlib \cite{dlib09} was sometimes unable to find landmarks in the ground truth videos. In those case the samples are ignored for the two metrics. 
We employ the SyncNet \cite{chung2017out} distance and confidence score towards evaluating lip synchronisation of generated video and audio. For those last two metrics, following the literature we simply feed the generated videos into the code provided by the author of \cite{chung2017out}.

\end{document}